\documentclass[conference]{IEEEtran}
\usepackage{cite}
\usepackage{amsmath,amssymb,amsfonts}
\usepackage{algorithmic}
\usepackage{algorithm} 
\usepackage{graphicx}
\usepackage{textcomp}
\usepackage{booktabs}
\usepackage{tabularx}
\usepackage{multirow}
\usepackage{algorithm}
\usepackage{amssymb}
\usepackage{hyperref}
\usepackage{subcaption} 
\usepackage[T1]{fontenc}
\usepackage{array}
\usepackage{graphicx}
\usepackage[utf8]{inputenc}
\usepackage[export]{adjustbox}
\usepackage{tcolorbox}
\usepackage{ulem}
\usepackage{float}

\def\BibTeX{{\rm B\kern-.05em{\sc i\kern-.025em b}\kern-.08em
    T\kern-.1667em\lower.7ex\hbox{E}\kern-.125emX}}

\begin{document}
\title{FedStain: Modeling Higher-Order Stain Statistics for Federated Domain Generalization in Computational Pathology}
\author{
    \IEEEauthorblockN{
        Fengyi Zhang\textsuperscript{1, \#}, 
        Junya Zhang\textsuperscript{2, \#}, 
        Wenzhuo Sun\textsuperscript{3, *}
    }
    
    \vspace{1em} 
    
    \IEEEauthorblockA{
        \textsuperscript{1}School of Electronic Science and Technology, Hainan University, Haikou, China, 570228
    }
    \IEEEauthorblockA{
        \textsuperscript{2}School of Computer Science and Technology, Xidian University, Xi'an, China, 710126
    }
    \IEEEauthorblockA{
        \textsuperscript{3}Xiangjiang College of Elite Engineers, Hunan University, Changsha, China, 410082
    }
    
    \IEEEauthorblockA{
        \textsuperscript{*}Corresponding author: sunwenzhuo@hnu.edu.cn
    }
    \IEEEauthorblockA{
        \textsuperscript{\#}These authors contributed equally.
    }
}

\maketitle

\begin{abstract}
Robust whole-slide image (WSI) analysis under stringent data-governance constraints remains highly challenging due to substantial stain heterogeneity across institutions. While domain generalization (DG) techniques aim to mitigate such cross-center distribution shifts, most require centralized access to multi-site data and therefore fundamentally conflict with real-world privacy regulations. Federated learning (FedL) offers a decentralized alternative; however, existing FedL and federated DG (FedDG) approaches rely almost exclusively on low-order statistics (e.g., means and variances), implicitly assuming Gaussian-like stain distributions. In contrast, real-world histopathological staining processes often produce asymmetric and heavy-tailed color distributions due to biochemical diffusion and scanner nonlinearity. As a result, current methods do not explicitly model the higher-order, non-Gaussian characteristics that dominate real-world stain variability. To address this limitation, we propose FedStain, a stain-aware FedDG framework that explicitly incorporates higher-order stain moments—skewness and kurtosis—as compact statistical descriptors exchanged during federated optimization. These descriptors introduce no pixel-level data transmission, thereby preserving strict privacy guarantees and communication efficiency, while enabling the global model to capture stain-style variability that low-order statistics inherently miss. FedStain further adopts a contrastive, cross-site parameter aggregation strategy to promote stain-invariant, institution-agnostic representations without relaxing any data-governance constraints. Extensive experiments on Camelyon17 and our newly constructed MvMidog-Fed benchmark demonstrate that FedStain yields consistent and substantial improvements, outperforming state-of-the-art FedL, DG, and FedDG baselines by up to +3.9\% absolute accuracy. To our knowledge, FedStain is the first federated DG approach to explicitly model higher-order stain statistics, marking a significant step toward robust cross-institutional deployment in gigapixel-scale computational pathology.

\end{abstract}

\section{Introduction}
\label{sec:introduction}
Artificial intelligence (AI) has become a foundational technology in modern medical diagnosis, with extensive empirical evidence demonstrating its value in enhancing diagnostic accuracy, improving service efficiency, broadening clinical accessibility, and strengthening the standardization of medical decision-making \cite{b1,b2,b51,b52,b5,b54,b55,b75}. In radiological screening, McKinney et al. reported that AI systems not only surpass average radiologists in breast cancer prediction but also reduce the double-reading workload by approximately 88\%, markedly improving operational efficiency \cite{b13}. In pathology, results from the CAMELYON16 challenge showed that deep learning models achieve pathologist-level performance in lymph node metastasis detection\cite{b14}. In resource-limited regions, mobile–AI screening solutions have been deployed to support cervical cancer detection, significantly lowering diagnostic barriers \cite{b15}. Furthermore, AI-driven clinical decision support systems have demonstrated potential in improving outcomes in critical-care scenarios such as sepsis management \cite{b16}. Collectively, these advances indicate that AI is reshaping diagnostic workflows on multiple fronts and is increasingly central to addressing global disparities in diagnostic quality and medical resource distribution.

However, the development of high-performance medical AI models—particularly for image analysis—critically depends on access to large-scale, diverse datasets. This requirement conflicts with the realities of clinical data governance: stringent regulations such as HIPAA and the GDPR, coupled with concerns over patient-privacy risks, lead to highly siloed medical data\cite{b3,b4}. A 2023 report from the China Academy of Information and Communications Technology (CAICT)\cite{b5} estimates that the cross-institutional data sharing rate of domestic medical institutions is less than 10\%, restricting AI training to narrow, institution-specific distributions and substantially impairing generalization in real-world clinical deployment.

In histopathological whole-slide image (WSI) analysis—an essential pillar of computational pathology—such fragmentation further exacerbates domain heterogeneity. WSIs produced by different institutions often exhibit pronounced stain variability due to differences in staining reagents, tissue preparation protocols, and scanner hardware\cite{b5,b6,b74}. Even minor variations in staining or scanning procedures can significantly alter tissue color distributions, as shown by Bejnordi et al. \cite{b17}, affecting the detection of key diagnostic cues such as tumor boundaries and cellular morphology. This stain-induced domain heterogeneity undermines cross-institutional diagnostic consistency and severely limits the reliability of computer-aided diagnosis (CAD) systems trained on data from a single source. 

These stain differences create a substantial domain shift: models trained on one domain frequently display degraded performance when applied to unseen institutional domains. DG methods aim to mitigate such shifts by learning domain-invariant representations from multiple source domains \cite{b7,b8,b9,b49,b50,b56,b57,b58,b59,b60,b61,b62}. Yet, most DG frameworks require centralized access to multi-domain data—an assumption that is often incompatible with prevailing privacy regulations governing clinical WSI data. Since pathology images inherently contain patient information, centralized aggregation may introduce substantial privacy risks and regulatory conflicts, hindering cross-institutional model development.

In contrast, the FedStain framework developed in this study leverages FedL to enable decentralized model training: institutions train models locally and share only parameter updates, eliminating the need to transmit raw WSIs and ensuring adherence to privacy regulations. Its stain-aware design further adapts the learning process to the characteristics of WSI data, enabling it to overcome key limitations of traditional DG approaches in privacy-sensitive, cross-institutional settings.

FedL has emerged as a promising paradigm for reconciling privacy constraints with collaborative learning. By iteratively aggregating locally trained model parameters on a central server, FedL enables distributed optimization without exposing sensitive data\cite{b10,b11}. This property makes it a natural foundation for FedStain. Yet, while FedL preserves privacy, it does not explicitly address domain heterogeneity, and models trained under standard FedL often exhibit weak performance on unseen institutions—particularly when stain variability is substantial\cite{b18,b76}. FedStain addresses this limitation by integrating DG principles into FedL, designing a stain-aware, domain-invariant learning mechanism that explicitly models cross-domain commonalities in WSIs and substantially enhances generalization across unseen institutions.

Recent developments in Federated Domain Generalization (FedDG) combine the strengths of FedL and DG to address the dual problems of privacy constraints and generalization. Representative methods such as FedCCRL, FedDG-ELCFS, and FedAlign utilize techniques including contrastive learning, frequency-space interpolation, and cluster-aware optimization to improve cross-client representation alignment\cite{b12,b19,b20}. However, these frameworks are predominantly designed for natural images and overlook the stain-specific factors that uniquely characterize histopathological data. A critical yet underexplored issue is that stain distributions in histopathology are inherently non-Gaussian, arising from complex biochemical diffusion, dye–tissue interactions, and scanner-induced nonlinearities. These processes naturally induce asymmetric and heavy-tailed color distributions. As a result, they violate the implicit Gaussian assumptions underlying most existing FedL, DG, and FedDG methods, which primarily align low-order statistics such as means and variances. As a result, general FedDG approaches are typically insufficient to capture the higher-order statistical dependencies that dominate real-world stain variability, leading to limited robustness under severe stain-induced domain shifts.

FedStain explicitly addresses this gap by incorporating the RandStainNA stain augmentation mechanism into the FedDG framework. Through adaptive simulation and refinement of stain-style variation, FedStain captures the dynamic and non-Gaussian statistical properties of stain distributions, enabling more accurate modeling of stain heterogeneity. This design substantially enhances the model’s ability to represent cross-institutional stain variability and significantly improves diagnostic robustness compared with general FedDG methods.

In computational pathology, extensive research has been devoted to mitigating stain variation, as effective stain modeling is essential for robust generalization. Methods such as STRAP apply medically irrelevant style transfer to disentangle domain-specific textures from semantic features\cite{b21}, while StainGAN employs CycleGAN-based stain–style transfer to standardize color distributions across scanners \cite{b22}. However, although augmentation techniques like RandStainNA broaden training distribution coverage, they typically rely on low-order statistics (e.g., mean, variance) and therefore fail to capture the inherently non-Gaussian properties of stain distributions—such as asymmetric diffusion patterns—and exhibit limited robustness under significant cross-institutional stain heterogeneity.

FedStain directly addresses these limitations by incorporating higher-order staining statistics (skewness, kurtosis) into the stain augmentation process. These descriptors accurately characterize the asymmetric and heavy-tailed properties of real-world stain distributions and, when integrated with an enhanced RandStainNA pipeline, significantly improve the expressiveness and robustness of stain modeling.

\subsection{Research Gap}

Despite progress in FedL, DG, FedDG, and stain normalization techniques, existing approaches have notable limitations: FedDG methods often treat clients as generic visual domains, overlooking the stain-specific non-Gaussian statistical properties that drive cross-site variability in histopathology \cite{b72}. DG frameworks assume centralized data and cannot model the complex stain distributions that dominate WSI appearance. Stain augmentation methods rely primarily on low-order statistics and remain disconnected from federated optimization \cite{b73}.

\vspace{\baselineskip}

Thus, no existing framework jointly models stain-specific variability and multi-site heterogeneity under real-world privacy constraints.

\subsection{Our Approach}

To address this gap, we propose FedStain, a stain-aware federated domain generalization framework that integrates higher-order stain statistics—skewness and kurtosis—into the federated optimization pipeline. 

\vspace{\baselineskip}

Unlike existing FedDG methods that either ignore stain characteristics or rely solely on low-order statistics (e.g., mean and variance), FedStain explicitly leverages higher-order stain moments to model non-Gaussian stain distributions across clients in a privacy-preserving manner.These compact descriptors capture complex non-Gaussian stain behaviors while requiring no pixel-level data sharing, ensuring compliance with privacy regulations. FedStain further incorporates a contrastive cross-site parameter aggregation strategy that aligns high-level semantic representations while preserving relevant stain-style characteristics, enabling robust, stain-invariant, institution-agnostic model generalization.

In addition, we construct MvMidog-Fed, a federated adaptation of the MIDOG dataset that incorporates cross-institutional domain partitioning and stain-feature annotation to establish a dedicated benchmark for stain-aware federated generalization.

The main contributions of this study can be summarized as follows:  
\begin{itemize}
    \item Stain-Aware Federated domain generalization Framework: We propose the first stain-aware federated domain generalization framework (FedStain), which integrates higher-order staining statistics (skewness-kurtosis) into the federated domain generalization paradigm to achieve robust domain-invariant learning in federated scenarios.  
    \item Construction of Federated Pathological Dataset: A standardized data processing pipeline is developed for the MIDOG dataset, providing support for reproducible and privacy-compliant federated domain generalization research in digital pathology.  
    \item Comprehensive Experimental Validation: Extensive experiments are conducted on the Camelyon17 dataset and the improved MIDOG dataset. The results show that FedStain consistently outperforms current mainstream benchmark models (FedAvg, FedProx, FedCCRL, FedAlign), verifying its effectiveness and scalability in practical cross-institutional diagnosis. 
\end{itemize}

\section{Related Work}

This section reviews prior research on the technical foundations of the proposed framework, including the core mechanisms of federated learning, the theoretical principles of domain generalization, algorithmic designs for federated domain generalization, and stain normalization, augmentation techniques in computational pathology. Emphasis is placed on the evolution and technical logic of pivotal methods.

\subsection{Federated Learning (FedL)}

FedL is a decentralized collaborative learning paradigm that enables privacy-preserving model training across distributed data silos without sharing raw data. In its standard server–client architecture, clients train models locally on institutional data and communicate only model parameters for aggregation on a central server, ensuring compliance with regulations such as HIPAA and GDPR\cite{b3,b4}.

FedAvg established the foundational framework of local training combined with weighted global aggregation, balancing learning performance and communication efficiency\cite{b10}. Subsequent studies such as FedProx, FedNova, and SCAFFOLD addressed challenges including client drift, data imbalance, and training variance\cite{b11,b25,b26}, improving stability and scalability.

In medical imaging—where data sharing is tightly restricted—FedL has facilitated multi-center collaboration across radiology, ophthalmology, and histopathology. However, traditional FedL lacks explicit mechanisms to systematically address cross-client domain heterogeneity.\cite{b42} As a result, models trained under FedL often exhibit degraded performance on unseen institutions with distinct imaging characteristics, motivating the development of FedDG.

FedStain builds on this foundation by incorporating DG principles and explicitly modeling stain-specific heterogeneity through higher-order staining statistics, enabling effective adaptation to cross-institutional stain variability while retaining the privacy-preserving advantages of FedL.

\subsection{Domain Generalization (DG)}

DG aims to learn representations that generalize to unseen domains without accessing target-domain data during training\cite{b7,b8,b9}, and its approaches mainly fall into three categories: the first is data-level methods, which expand the diversity of training distributions through data augmentation or synthetic domain creation (e.g., MixStyle, DomainBed) \cite{b9,b27}; the second is feature-level methods, which achieve cross-domain alignment by leveraging adversarial learning (e.g., DANN), MMD-based adaptation, or contrastive learning\cite{b7,b28}; the third is model-level methods, which enhance robustness via episodic meta-learning or regularization strategies (such as MLDG, MASF, and EISNet)\cite{b8,b29,b30}.

While these methods are effective in natural image tasks, they face challenges in medical imaging\cite{b35,b36,b41,b45,b46,b47,b48}, where domain shifts arise from hardware variations, scanning protocols, and tissue preparation processes—factors that induce complex, non-Gaussian distribution changes. DG methods also assume centralized data access, conflicting with stringent privacy requirements in clinical WSI analysis.

FedStain resolves these issues by integrating decentralized training with a stain-aware mechanism that incorporates higher-order staining statistics, enabling more accurate modeling of pathology-specific domain shifts.

\subsection{Federated Domain Generalization (FedDG)}

FedDG\cite{b12,b18,b19,b63,b64,b65} combines the privacy-preserving properties of FedL with the robustness objectives of DG, enabling cross-domain generalization under decentralized data conditions. Clients are treated as independent domains, and federated aggregation aims to produce a model capable of generalizing to unseen domain distributions.

Recent advancements include FedDG, which introduces episodic learning in the frequency domain for segmentation tasks\cite{b19}; FedCCRL, which employs cross-client contrastive learning to enhance feature consistency\cite{b12}; and methods such as FedBN and MOON, which mitigate domain interference via batch normalization decoupling and momentum contrast, respectively\cite{b18}. These studies form a technical foundation for generalization under non-IID conditions.

However, these approaches primarily target semantic or frequency-based shifts in natural images and do not explicitly model the statistical shape of stain distributions, leaving an important gap for histopathology-specific federated generalization.

FedStain addresses this gap by integrating stain augmentation (via RandStainNA) with federated optimization and embedding higher-order staining statistics to explicitly capture the non-Gaussian properties of stain distributions. This enables effective stain-aware domain-invariant learning in decentralized training environments.

\subsection{Stain Normalization and Augmentation Techniques in Computational Pathology}

Stain variation is the primary source of domain shift in computational pathology, driven by differences in H\&E staining intensity, scanner illumination, and slide preparation. Existing approaches fall into two major categories: Early approaches relied on color deconvolution and statistical template matching\cite{b31,b32}. Bejnordi et al. introduced stain-specific normalization techniques, while Reinhard-based methods matched global color statistics. Later, generative models such as StainGAN leveraged CycleGAN-based style transfer to achieve stain normalization across scanners\cite{b22}.

However, template-based methods suffer from sensitivity to reference choices, and GAN-based methods risk introducing unrealistic artifacts or compromising fine-grained tissue structure.

Stain augmentation\cite{b31,b32,b66,b67,b68,b69,b70,b71} aims to improve robustness by simulating stain variability. STRAP applies medically irrelevant style transfer to achieve domain-agnostic feature learning, while RandStainNA unifies stain normalization and augmentation by stochastically perturbing mean and variance parameters\cite{b21,b23}. RandStainNA++ further enhances robustness by incorporating foreground–background separation and self-distillation\cite{b33}.

Despite their effectiveness, these methods rely primarily on low-order statistical modeling,which is often insufficient to fully represent the asymmetric, heavy-tailed nature of real-world stain distributions.
FedStain advances this research direction by introducing higher-order staining statistics (skewness and kurtosis) to capture complex non-Gaussian characteristics, and by embedding this enhanced stain-aware mechanism into a federated DG framework. This design enables privacy-compliant, stain-aware augmentation and model training across institutions, addressing limitations of prior stain modeling and enabling robust cross-institutional deployment.


\section{Method}

We propose a novel Federated Domain Generalization (FedDG) framework, named \textbf{FedStain}, specifically designed to address cross-domain generalization challenges under Federated Learning (FedL) for histopathological image analysis. By integrating stain-aware data augmentation within a federated architecture, FedStain mitigates domain shift while preserving data privacy, enabling the model to learn domain-invariant representations from distributed stain-heterogeneous histopathology images.

\subsection{Preliminaries}
\label{sec:preliminaries}

We consider a FedL setup with $K$ clients. Each client $i \in \{1, \dots, K\}$ holds a local dataset $D_i = \{(x_j, y_j)\}$, where samples $(x, y)$ follow a domain-specific distribution $P_i(x, y)$. Here, $x$ denotes a histopathological image and $y$ its label. The target domain $P_t$ corresponds to an unseen institution, satisfying $P_t \neq P_i$ for all $i$ \cite{b20}.  

Two core challenges arise in medical histopathology: 1) stringent privacy requirements (e.g., GDPR, HIPAA) prohibit sharing raw patient data, and 2) inter-institutional stain heterogeneity leads to severe degradation in model generalization \cite{b1}. For example, slides from different institutions often exhibit systematic variations in hematoxylin and eosin intensity distributions. 

\begin{figure}[H] 
    \centering 
    \includegraphics[width=\linewidth]{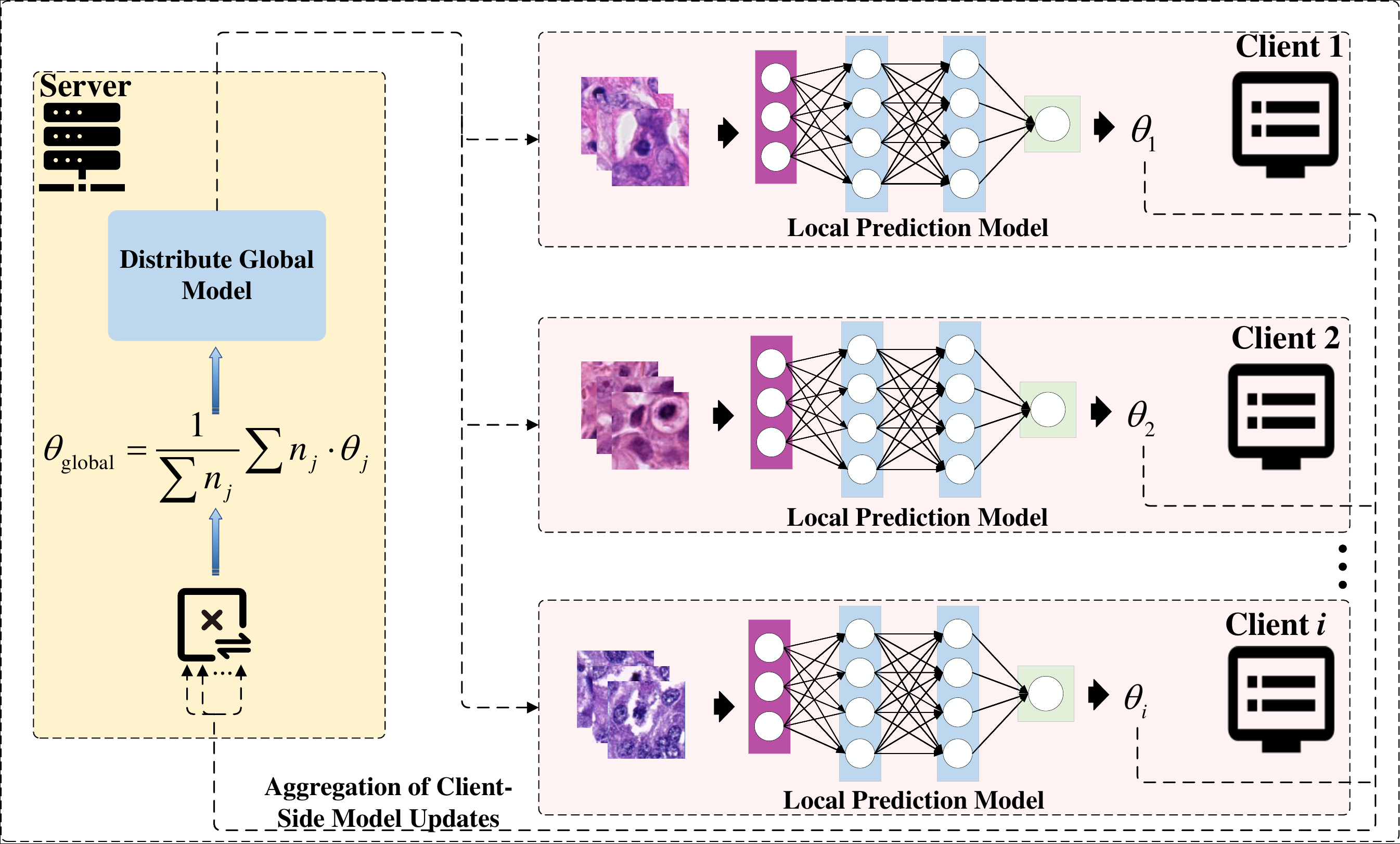} 
    \caption{Overview of standard FedAvg. It establishes a privacy-preserving client–server training paradigm, but does not explicitly address inter-client stain heterogeneity.}
    \label{fig:fedavg_fixed}
\end{figure}

\subsection{Unified Design Motivation}
\label{sec:Unified Design Motivation}

Domain shifts in federated histopathology could arise from two sources: 1) raw-level stain variability and 2) higher-level feature representations and tissue morphology discrepancies. These sources arise at different levels along the image-to-feature hierarchy and therefore require level-specific mitigation strategies.

\paragraph{Level 1: Raw-Level Stain Diversification.}
Variations in staining protocols, reagents, and scanner profiles constitute a primary source of domain shift in histopathological images. Level-1 interventions enrich stain diversity directly in the pixel space, broadening the range of stain conditions during local training and reducing the risk of site-specific bias.

\paragraph{Level-2: Feature-Level Morphology and Style Perturbation.}  
After raw-level stain inconsistencies are mitigated, remaining domain shifts manifest at the feature level. Level-2 introduces:  
\begin{itemize}
    \item \textbf{Morphology-oriented perturbations:} controlled transformations that alter geometric and structural attributes while preserving semantic content, preventing reliance on local morphological idiosyncrasies.
    \item \textbf{Style-oriented feature perturbations:} adjustments of statistical properties of intermediate representations to capture higher-order stain-related variations beyond the pixel space.
\end{itemize}  
This hierarchical design enables progressive mitigation of domain shift by first reducing raw stain variability and subsequently addressing higher-level discrepancies in feature representations and tissue morphology. By mirroring the hierarchical origins of domain shifts, each level is handled using its most appropriate intervention: Level-1 targets raw stain variations, while Level-2 focuses on feature- and morphology-level distribution differences.

From the distributional perspective, Level-1 augmentation expands the support of the input distribution, while Level-2 enforces invariance in the latent feature space, jointly approximating a minimax generalization objective under unseen stain shifts.

\subsection{Level-1: Raw Stain Diversification via RandStain}
\label{sec:randstain}

Domain shift originates at the raw pixel level, where variations in staining protocols, reagents, and scanner profiles produce heterogeneous color distributions. To mitigate this source of shift, we introduce \textbf{RandStain}, a raw-level stain diversification mechanism that operates prior to any feature-level augmentation. RandStain simulates realistic, previously unseen stain patterns based on statistical descriptors collected across clients in the federated environment \cite{b33}.

The augmentation process follows a three-step pipeline: \emph{statistic sampling $\rightarrow$ parameter calculation $\rightarrow$ stain reconstruction}. For each channel $ch$ of input image $x$, the augmented image $x'$ is computed as:
\begin{equation}
x'_{ch} = (x_{ch} - \mu_{ch}) \cdot \sigma'_{ch} + \mu'_{ch},
\label{eq:randstain}
\end{equation}
where $\mu'_{ch}$ and $\sigma'_{ch}$ are sampled from the distribution of training statistics to ensure realistic, diagnostically plausible variations. By enriching the raw stain space, Level-1 establishes a foundational domain envelope for subsequent Level-2 feature-level interventions. Through channel-wise statistic sampling across clients, RandStain introduces realistic variations while preserving morphological structures critical for diagnosis. 

\subsection{Level-2: Domain Generalization via AugMix and MixStyle}
\label{sec:domain_generalization}

To further address domain diversity, Level-2 jointly leverages \textbf{AugMix} and \textbf{MixStyle}. AugMix generates morphology-related perturbations, while MixStyle adapts feature-level staining statistics. Together, they form a comprehensive Level-2 mechanism for high-level domain shift mitigation.

\subsubsection{AugMix: Morphology-Level Perturbation}
\label{sec:augmix}
Within Level-2, AugMix introduces multi-chain stochastic perturbations—flipping, rotation, scaling, and composition—to expand feature distribution diversity. This improves robustness to minor acquisition noise and prevents overfitting to local morphological patterns \cite{b34}. However, because AugMix does not explicitly alter stain-related feature statistics, residual stain-induced shifts remain, motivating the complementary use of MixStyle.

\begin{figure*} 
    \centering
    \includegraphics[width=0.95\textwidth]{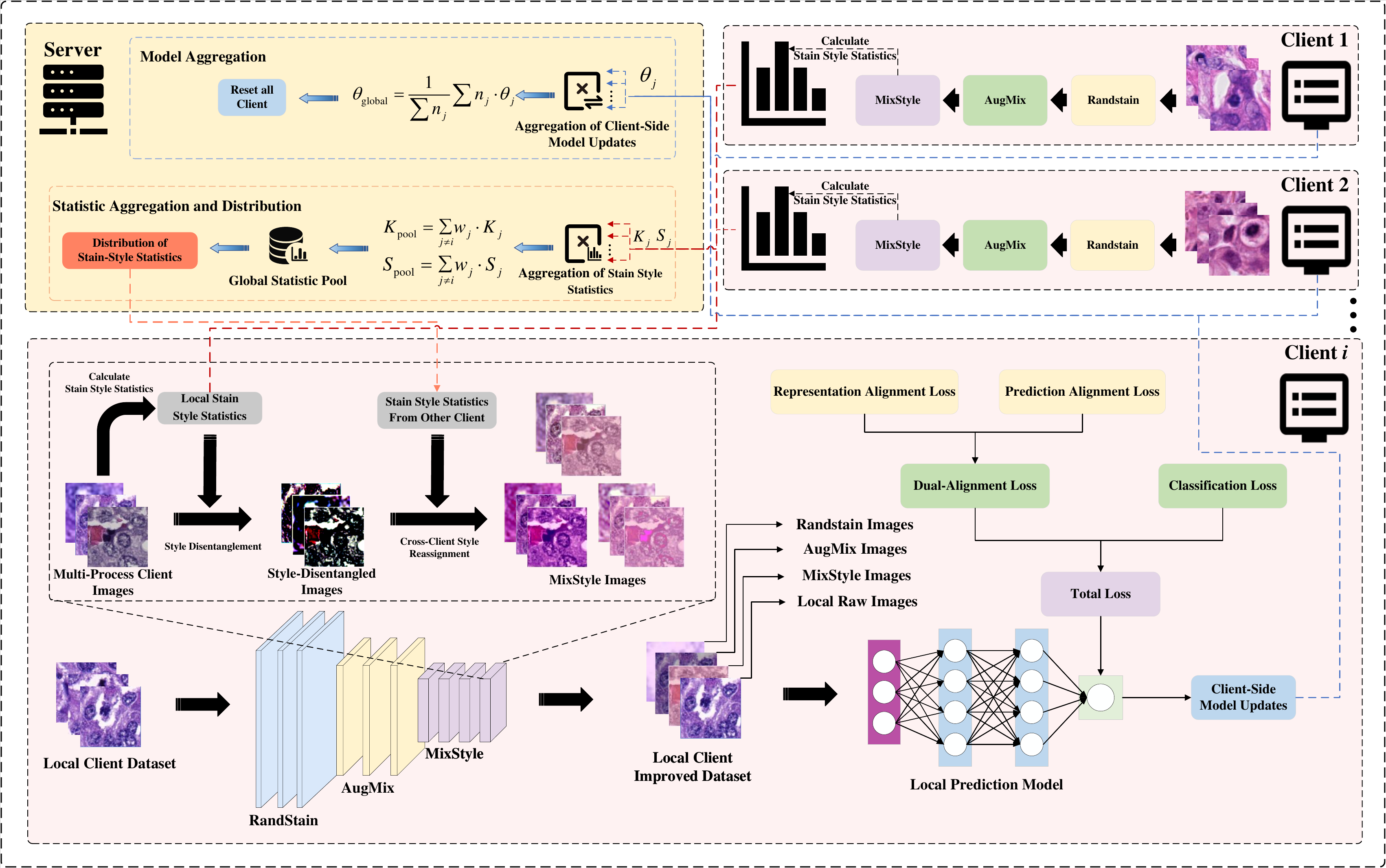} 
    \caption{Overview of FedStain. Each client calculates local stain style statistics and transmits them, along with updated local model parameters, to the server. The server aggregates the received stain style statistics to construct a global statistic pool, then redistributes the pool excluding each client’s own statistics back to the corresponding client; meanwhile, the server performs aggregation of client-side model updates to generate a global model. During the local training phase, RandStain, AugMix, and MixStyle are sequentially applied for stain diversity expansion, morphological perturbation, and cross-client style reassignment. Additionally, the dual-alignment loss (including representation alignment loss and prediction alignment loss) and classification loss are jointly utilized to constrain the model, ensuring it focuses on learning domain-invariant features while mitigating the impacts of stain heterogeneity and non-IID data distribution in federated scenarios. }
    \label{fig:fedstain_bottom}
\end{figure*}

\subsubsection{MixStyle: Feature-Level Stain Adaptation}
\label{sec:mixstyle}
While AugMix increases robustness to morphological variations, MixStyle specifically addresses feature-level stain distribution shifts, making the Level-2 augmentation comprehensive. 
MixStyle complements AugMix by adapting intermediate feature statistics to match inter-client stain distributions. While mean and standard deviation capture basic style shifts, complex stain patterns often exhibit skewed or heavy-tailed distributions. For instance, in the MvMIDOG dataset, slices from different scanners show distinct skewness and kurtosis patterns \cite{b24}. We therefore incorporate higher-order statistics—\textbf{skewness ($S$)} and \textbf{kurtosis ($K$)} \cite{b27,b32}—to better capture these differences. For channel $c$:
\begin{equation}
S_c = \frac{E[(x_c - \mu_c)^3]}{(\sigma_c^2)^{3/2}}, \quad
K_c = \frac{E[(x_c - \mu_c)^4]}{(\sigma_c^2)^2},
\label{eq:skewness_kurtosis_channel}
\end{equation}
where $\mu_c$ and $\sigma_c$ are the mean and standard deviation of channel $c$. MixStyle then performs cross-client statistic sharing and local style fusion, enabling adaptive handling of stain heterogeneity. Notably, skewness and kurtosis constitute the lowest-order moments that characterize distributional asymmetry and tail behavior beyond Gaussian assumptions, making them particularly suitable for communication-efficient federated settings. Combined with AugMix, Level-2 addresses both morphology- and stain-level feature domain shifts.

For illustration, the following pseudocode Algorithm 1 shows the MixStyle implementation for cross-client stain adaptation 

\begin{algorithm}[h]
\caption{MixStyle with Higher-Order Statistics}
\label{alg:mixstyle}
\begin{algorithmic}[1] 
\REQUIRE Batch $X$, Parameter $\alpha$, Pools $S_{\text{pool}}^i, K_{\text{pool}}^i$
\ENSURE Stylized Batch $X_{\text{mixstyle}}$

\STATE Initialize empty list $\hat{X} = []$
\FOR{each sample $x \in X$}
    \STATE Calculate $S(x), K(x)$ Eq. \ref{eq:skewness_kurtosis_channel}
    \STATE Standardize $x$: $\tilde{x} = (x - \mu) / \sigma$
    \STATE Sample $(S', K')$ from $S_{\text{pool}}^i, K_{\text{pool}}^i$
    \STATE Sample $\lambda \sim \text{Beta}(\alpha, \alpha)$
    \STATE Compute mixed statistics:
        \STATE $\beta_{\text{mix}} \leftarrow \lambda \cdot S' + (1-\lambda) \cdot S(x)$
        \STATE $\gamma_{\text{mix}} \leftarrow \lambda \cdot K' + (1-\lambda) \cdot K(x)$
    \STATE $\hat{x} \leftarrow \gamma_{\text{mix}} \cdot \tilde{x} + \beta_{\text{mix}}$
    \STATE Append $\hat{x}$ to $\hat{X}$
\ENDFOR
\STATE $X_{\text{mixstyle}} = \text{cat}(\hat{X})$

\end{algorithmic}
\end{algorithm}
\begin{algorithm*}[t!]
\caption{The FedStain Framework}
\label{alg:fedstain}
\begin{algorithmic}[1] 
\REQUIRE Statistic upload ratio $r$, hyperparameters $\alpha, \beta$
\ENSURE Global model parameters $\theta^{T+1}$

\STATE \textbf{Server:} Initialize model $f = g \circ h$ ($h$: encoder, $g$: classifier), distribute $\theta^0$ to all clients.
\FOR{each communication round $t = 1, 2, \dots, T$}
    \STATE \COMMENT{Stage (a): Stain Statistic Collection}
    \FOR{each client $i$ (in parallel)}
        \STATE Calculate stain statistics (Skewness $S_i$, Kurtosis $K_i$) for $\lceil r \cdot n_i \rceil$ local samples Eq. \ref{eq:skewness_kurtosis_channel}
        \STATE Upload $S_i, K_i$ to the server.
    \ENDFOR
    \STATE \textbf{Server:} Aggregate statistics into global pools: $S_{pool} = \{S_i\}_{i=1}^K$, $K_{pool} = \{K_i\}_{i=1}^K$.
    \STATE \textbf{Server:} Distribute partial pools to client $i$: $S_{pool}^i = S_{pool} \setminus S_i$, $K_{pool}^i = K_{pool} \setminus K_i$.
    
    \STATE \COMMENT{Stage (b): Local Training}
    \FOR{each client $i$ (in parallel)}
        \FOR{each local epoch $e = 1, 2, \dots, E$}
            \FOR{each batch $X \in D_i$}
                \STATE $X^{stain} \leftarrow \text{RandStain}(X, S_{pool}^i, K_{pool}^i)$  Eq. \ref{eq:randstain}
                \STATE $X^{(1)} \leftarrow \text{MixStyle}(X, S_{pool}^i, K_{pool}^i) \circ \text{AugMix}(X)$ (See Alg. \ref{alg:mixstyle})
                \STATE $X^{(2)} \leftarrow \text{MixStyle}(X, S_{pool}^i, K_{pool}^i) \circ \text{AugMix}(X)$
                \STATE Extract features: $Z = h(X)$, $Z^{(1)} = h(X^{(1)})$, $Z^{(2)} = h(X^{(2)})$, $Z^{stain} = h(X^{stain})$
                \STATE Extract predictions: $\hat{Y} = g(Z)$, $\hat{Y}^{(1)} = g(Z^{(1)})$, $\hat{Y}^{(2)} = g(Z^{(2)})$, $\hat{Y}^{stain} = g(Z^{stain})$
                \STATE Calculate $L_{CLS}$ Eq. \ref{eq:supcon_loss},  $L_{RA}$ Eq. \ref{eq:RA_loss}, $L_{JS}$ Eq. \ref{eq:JS_loss}, 
                \STATE Compute total loss $L = L_{CLS} + \alpha L_{RA} + \beta L_{JS}$ Eq. \ref{eq:Total_loss}
                \STATE Update local parameters $\theta_i^t$ via Adam optimizer.
            \ENDFOR
        \ENDFOR
        \STATE Upload updated $\theta_i^t$ to server.
    \ENDFOR
    
    \STATE \COMMENT{Stage (c): Global Aggregation}
    \STATE \textbf{Server:} Update global parameters via FedAvg: $\theta^{t+1} \leftarrow \frac{1}{N} \sum_{i=1}^K n_i \cdot \theta_i^t$
    \STATE \textbf{Server:} Distribute $\theta^{t+1}$ to all clients.
\ENDFOR
\STATE \textbf{Return} final global parameters $\theta^{T+1}$

\end{algorithmic}
\end{algorithm*}
\subsection{Loss Function Construction}
\label{sec:loss}

To synergistically optimize for both model classification accuracy and cross-augmentation prediction consistency, our loss function employs a weighted fusion structure: a  \textbf{Classification Loss($L_{CLS}$)} and a  \textbf{Dual-Alignment Loss}. 
Level-1 augmentation (RandStain) primarily contributes to feature-level alignment, while Level-2 augmentations (AugMix+MixStyle) provide additional constraints for both feature and prediction-level consistency. 
The Dual-Alignment Loss is therefore composed of a  \textbf{Representation Alignment Loss ($L_{RA}$)} and a \textbf{Prediction Alignment Loss ($L_{JS}$)}.

\subsubsection{Classification Loss and Representation Alignment Loss}

To enhance feature discriminability across domains, we adopt the Supervised Contrastive Loss ($L_{SupCon}$) \cite{b37}, which constructs positive (same-class) and negative (different-class) pairs based on sample labels:

\begin{multline}
L_{\text{SupCon}} = \sum_{i \in I} \frac{-1}{|P(i)|} \sum_{p \in P(i)} 
\log\left( \frac{e^{\text{s}(z_i, z_p)/\tau}}{\sum_{a \in A(i)} e^{\text{s}(z_i, z_a)/\tau}} \right)
\label{eq:supcon_loss}
\end{multline}

\noindent where $I$ is the index set of all samples, $P(i) = \{p \in A(i) : y_p = y_i\}$ is the set of positive samples (samples with the same label as $i$), $A(i) = I \setminus \{i\}$ is the set of all samples excluding $i$, $\text{s}(\cdot, \cdot)$ denotes cosine similarity, and $\tau$ is a temperature parameter.

Building on this, we define the Representation Alignment Loss ($L_{RA}$) as the average supervised contrastive loss between the original features and augmented features from both Level-1 (RandStain) and Level-2 (AugMix+MixStyle):

\begin{equation}
L_{RA} = \frac{1}{3} \sum_{X \in \{Z^{stain}, Z^{(1)}, Z^{(2)}\}} L_{\text{SupCon}}(Z, X)
\label{eq:RA_loss}
\end{equation}
\subsubsection{Prediction Alignment Loss}
Constraints at the feature layer alone do not guarantee predictive stability. If the features of original and augmented samples are similar but their prediction distributions diverge, the model may still be perturbed by stain styles or noise. To maintain stable prediction outputs across augmentations, we employ the Jensen-Shannon (JS) Divergence \cite{b39}to define the Prediction Alignment Loss ($L_{JS}$).

JS divergence is a symmetric and bounded (0-1) variant of the Kullback-Leibler (KL) divergence, offering a more balanced measure of the difference between multiple probability distributions. The core logic is to first compute the average prediction distribution $\bar{Y}$ across all sample versions (original and augmentations), and then measure the KL divergence of each sample's prediction distribution relative to this average:

\begin{equation}
\begin{split}
L_{JS}(\hat{Y}_1, \dots, \hat{Y}_M) = \frac{1}{M} \sum_{i=1}^M D_{KL}(\hat{Y}_i || \bar{Y}), \\
\end{split}
\label{eq:JS_loss}
\end{equation}
\noindent where $\{\hat{Y}_i\}_{i=1}^M$ represents the set of prediction results for all $M$ versions of the samples, and $D_{KL}$ denotes the KL divergence.

\subsubsection{Total Loss Function: Multi-Objective Weighted Fusion}

To balance the primary classification objective with the dual-alignment constraints \cite{b40}, we fuse the classification loss ($L_{CLS}$), representation alignment loss ($L_{RA}$), and prediction alignment loss ($L_{JS}$) using hyperparameter weights, forming the final total loss function:

\begin{equation}
L = L_{CLS} + \alpha L_{RA} + \beta L_{JS}
\label{eq:Total_loss}
\end{equation}
\noindent where $\alpha$ and $\beta$ are hyperparameters (determined via grid search) that adjust the contribution weights of the representation and prediction alignment losses, respectively. This ensures the model achieves an optimal balance between classification accuracy and cross-domain generalization.

\subsection{Model Overview}

Algorithm~\ref{alg:fedstain} summarizes the complete client–server collaborative workflow of the \textbf{FedStain} model. It outlines the end-to-end training procedure, from parameter initialization to the final global model aggregation, facilitating reproducibility of the proposed pipeline.

\section{THEORETICAL JUSTIFICATION}

Existing Federated Domain Generalization (FedDG) methods, such as FedCCRL, adopt mean and variance as the exchanged statistics in MixStyle-based feature perturbation. This design is primarily motivated by the constraints inherent to federated learning, namely strict privacy preservation and communication efficiency. As low-order global statistics, mean and variance provide a compact summary of channel-wise feature distributions while introducing negligible communication overhead and avoiding any transmission of pixel-level information. Under these conditions, they have proven effective for style mixing and cross-domain alignment in natural image tasks, where feature distributions are often assumed to be approximately Gaussian \cite{b12}.\par

However, such low-order statistics are insufficient to characterize domain shifts induced by histopathological staining variability, particularly in whole-slide images (WSIs). Mean–variance-based style modeling implicitly assumes that the underlying stain distributions are approximately Gaussian, under which the first two moments adequately describe the distribution shape. In contrast, real-world histopathological staining processes are governed by complex biochemical diffusion, dye–tissue interactions, and scanner-dependent nonlinear responses \cite{b17}. These factors give rise to distinctly non-Gaussian stain distributions, which commonly exhibit pronounced asymmetry and heavy-tailed behavior. In such cases, mean and variance capture only central tendency and dispersion, while failing to encode distributional shape, resulting in an incomplete and potentially misleading representation of stain-induced domain heterogeneity.\par
\begin{figure*} 
    \centering
    \includegraphics[width=0.95\textwidth]{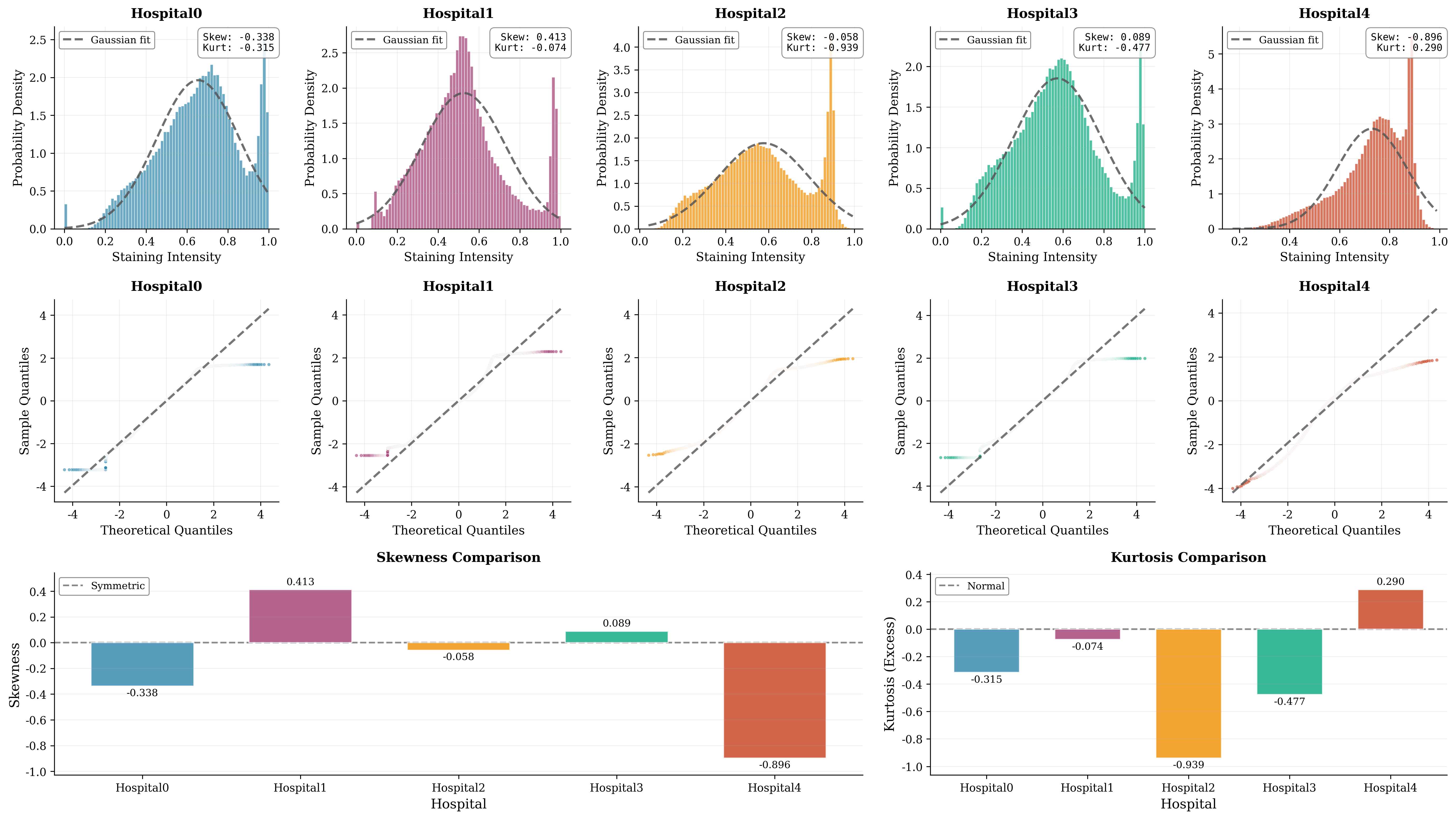} 
    \caption{Statistical characterization of staining intensity distributions across five hospital domains in the Camelyon17 dataset. The top row shows probability density histograms overlaid with Gaussian fits (dashed lines), illustrating clear deviations from Gaussianity across institutions. The second row presents Quantile–Quantile (Q–Q) plots, where systematic departures from the diagonal reveal pronounced non-Gaussian behavior, including asymmetry and heavy tails. The bottom row quantitatively compares skewness and excess kurtosis across hospitals. Notably, the distribution shapes vary substantially, ranging from a right-skewed pattern in Hospital 1 to a strongly left-skewed and leptokurtic distribution in Hospital 4. These results demonstrate that first- and second-order statistics alone cannot capture stain-induced distributional shape differences across domains, thereby motivating the incorporation of higher-order statistics in FedStain.}
    \label{fig:distributions}
\end{figure*}

To address this limitation, FedStain introduces skewness and kurtosis as higher-order stain statistics for cross-client exchange. The rationale behind this choice is supported from both empirical performance and distributional suitability perspectives.\par

From an empirical standpoint, ablation results (Table \ref{tab:mvmidog-features}) demonstrate that exchanging skewness and kurtosis consistently yields superior performance compared to mean–variance statistics and other alternative descriptors. On the MvMidog dataset, the skewness–kurtosis combination improves the average classification accuracy to 88.06\%, outperforming mean–variance statistics (84.66\%) as well as other candidates such as quantile-based or local dispersion measures. A particularly notable improvement is observed on Hospital 4 of the Camelyon17 dataset, which exhibits the most severe stain heterogeneity and non-Gaussian characteristics. In this setting, FedStain achieves an accuracy of 95.01\%, substantially surpassing FedCCRL (69.47\%) and FedAlign (79.34\%), highlighting the advantage of higher-order statistical modeling under extreme stain-induced domain shifts.\par

From the perspective of distribution adaptation, the visual analysis (Fig.3) further reveals the intrinsic differences in WSI staining distributions across varying institutions. Taking the Camelyon17 dataset as an example, the distribution shapes exhibit strong heterogeneity among sites: samples from Hospital 1 display a typical right-skewed distribution (Skewness $\approx 0.413$), with a tail extending towards higher intensities, whereas Hospital 4 exhibits a sharply contrasting pattern characterized by significant negative skewness (Skewness $\approx -0.896$) and a pronounced peak (Kurtosis $\approx 0.290$). These skewed and leptokurtic morphologies deviate markedly from the Gaussian assumption, under which mean-variance based descriptions become inadequate, as they inherently fail to distinguish between a flat, platykurtic distribution (e.g., Hospital 2, Kurtosis $\approx -0.939$) and a highly concentrated, leptokurtic one (e.g., Hospital 4). In contrast, FedStain incorporates skewness to quantify the direction of asymmetry and kurtosis to reflect the peakedness and tail weight. Together, they constitute the minimal yet expressive set of statistical descriptors for non-Gaussian descriptions, enabling a more complete capture of the key statistical features driving domain shifts in histopathology images.\par

Although more complex or higher-dimensional statistical descriptors could, in principle, capture additional aspects of stain distributions, their adoption must be carefully evaluated within the constraints of federated medical learning. First, communication efficiency is a critical consideration: frequent exchange of high-dimensional statistics would significantly increase communication costs and conflict with the lightweight design objectives of FedStain. In contrast, skewness and kurtosis require only third- and fourth-order moment computations, preserving a communication footprint of 2×C dimensions (C denotes the number of channels)—comparable to that of mean and variance. Second, privacy preservation is paramount in medical applications governed by regulations such as HIPAA and GDPR \cite{b3,b4}. Complex or spatially localized statistics may inadvertently encode fine-grained information that could facilitate data reconstruction attacks, whereas skewness and kurtosis remain global, non-invertible descriptors devoid of spatial cues. Finally, optimization stability and usability must be considered. Overly complex statistical perturbations may introduce excessive style distortion during local training, encouraging overfitting to non-essential distributional details and undermining the learning of domain-invariant semantic features—contrary to the clinical robustness objectives of FedStain.\par

In summary, the adoption of skewness and kurtosis in FedStain reflects a principled trade-off among representational adequacy, communication efficiency, privacy safety, and optimization stability. By explicitly modeling the non-Gaussian characteristics of stain distributions while respecting the practical constraints of federated medical learning, these higher-order statistics provide a compact yet expressive foundation for robust cross-institutional WSI analysis.

\section{Expirement}
To rigorously assess the effectiveness of the proposed FedStain framework, we conducted a comprehensive series of experiments on two medical image classification datasets: the widely used public dataset Camelyon17 and MvMidog, a newly curated dataset consisting of real-world histopathological images. Our approach was benchmarked against several representative federated domain generalization (FDG) methods, including FedAvg, FedProx, FedCCRL, and Strap, across different backbone architectures (ResNet18 and ResNet50).

\subsection{Datasets}
We evaluated the proposed method on two histopathological datasets characterized by substantial inter-domain heterogeneity.The Camelyon17 dataset comprises 455,954 images derived from five distinct hospitals, specifically designed for the domain generalization task. The dataset contains Whole-Slide Images (WSI) of Hematoxylin and Eosin (H\&E) stained lymph node sections, where the 'domains' are defined by the source hospitals. These hospitals introduce significant data distribution shifts due to differences in patient populations, slide staining protocols, and image acquisition systems.labeled into tumor and normal categories.To further evaluate model generalization in more diverse real-world settings. we constructed the MvMidog dataset based on the public MIDOG2025 dataset, which contains 503 original whole-slide images (WSIs). Using the tumor coordinate annotations provided by MIDOG2025, we extracted 128×128 patches centered on tumor regions as positive samples, and randomly cropped non-tumor regions as negative samples.For annotations located near the WSI boundaries, we implemented a boundary-checking mechanism: if the tumor area failed to fall fully within the field of view after three random attempts, we re-cropped patches using the image corners as new centers. Moreover, to mitigate potential overfitting caused by tumors consistently appearing at patch centers, tumor regions were randomly offset within each cropped image. To guarantee the integrity of our dataset, low-quality patches were filtered based on non-white area ratio, edge complexity, and color distribution—effectively removing overly homogeneous or structurally ambiguous samples. After rigorous data curation, we obtained a total of 52,571 high-quality image patches.\par
Following the scanner metadata provided in MIDOG2025, MvMidog was partitioned into four domains: 3D Histech, Aperio CS2, Hamamatsu S360, and Hamamatsu XR, each containing both tumor and normal samples.\par
For evaluation, we adopted a leave-one-domain-out (LODO) protocol. Specifically, in each iteration, one domain was treated as the target test domain, while the remaining domains served as source domains for federated training. This process was repeated until each domain had been used once as the target, ensuring a comprehensive assessment of cross-domain robustness.
\begin{table}[t]
\centering
\caption{Hyperparameters used in the came experiment}
\label{tab:single}
\begin{tabularx}{\linewidth}{c c X} 
\toprule
\textbf{Hyperparameter} & \textbf{Value} & \textbf{Clarification} \\ \midrule
n\_round & 3 & Number of times the server aggregates client models and updates the global model \\
batch\_size & 32 & Size of each batch for optimization algorithms \\
n\_epochs & 3 & Number of local training epochs before aggregation \\
learning\_rate & $1\times 10^{-5}$ & From $1 \times 10^{-4}$ to $2.5 \times 10^{-6}$, linear learning rate scheduling \\
alpha & 0.5 & Dirichlet parameter controlling the level of non-IID across clients \\
num\_clients\_per\_domain & 2 & Number of participating clients assigned to each domain \\ 
\bottomrule
\end{tabularx}
\end{table}

\begin{table}[t]
\centering
\caption{Hyperparameters used in the mvmidog experiment}
\label{tab:single1}
\begin{tabularx}{\linewidth}{c c X} 
\toprule
\textbf{Hyperparameter} & \textbf{Value} & \textbf{Clarification} \\ \midrule
n\_round & 3 & Number of times the server aggregates client models and updates the global model \\
batch\_size & 16 & Size of each batch for optimization algorithms \\
n\_epochs & 3 & Number of local training epochs before aggregation \\
learning\_rate & $1\times 10^{-4}$ & From $1 \times 10^{-4}$ to $2.5 \times 10^{-6}$, linear learning rate scheduling \\
alpha & 0.5 & Dirichlet parameter controlling the level of non-IID across clients \\
num\_clients\_per\_domain & 2 & Number of participating clients assigned to each domain \\ 
\bottomrule
\end{tabularx}
\end{table}
\subsection{Baselines and Implementation Details}
\begin{table*}[t]
\centering
\caption{Test accuracy on Camelyon17. In this set of experiments, we set the upload    ratio of the statistics r = 0.1. To ensure fairness, each algorithm is evaluated three times, and the final average is taken as the experimental result. }
\label{tab:camelyon}
\resizebox{\textwidth}{!}{
\begin{tabular}{lcccccc|cccccc}
\hline
& \multicolumn{12}{c}{\textbf{Camelyon17}}\\ 
& \multicolumn{6}{c|}{\textbf{ResNet18}} & \multicolumn{6}{c}{\textbf{ResNet50}} \\ 
& Hosp0 & Hosp1 & Hosp2 & Hosp3 & Hosp4 & Avg. & Hosp0 & Hosp1 & Hosp2 & Hosp3 & Hosp4 & Avg. \\ \hline
FedAvg & 87.37 & 70.07 & 66.44 & 59.52 & 84.64 & 73.61 & 78.96 & 80.20 & 56.73 & 56.10 & 64.06 & 67.21 \\
FedProx & 86.29 & 83.79 & 63.90 & 57.01 & 72.07 & 72.61 & 90.67 & 83.04 & 86.86 & 87.55 & 89.81 & 87.59 \\
FedCCRL & 94.89 & 92.26 & 93.61 & \textbf{95.82} & 73.55 & 90.56 & 94.93 & 92.50 & \textbf{95.81} & 93.27 & 69.47 & 89.20 \\
FedAlign & \textbf{96.44} & \textbf{93.91} & 94.05 & 94.79 & 70.00 & 89.84 & 95.88 & 91.42 & 94.47 & 93.90 & 79.34 & 91.00 \\
Strap & 87.78 & 84.53 & 90.86 & 90.53 & 88.54 & 88.49 & 89.61 & 87.78 & 91.89 & 91.56 & 90.00 & 90.17 \\
FedStain (Ours)& 95.25 & 89.16 & \textbf{94.43} & 89.93 & \textbf{93.06} & \textbf{92.37} & \textbf{96.90} & \textbf{93.42} & 94.21 & \textbf{94.80} & \textbf{95.01} & \textbf{94.87} \\ \hline
\end{tabular}
}
\end{table*}
\begin{table*}[t] 
\centering 
\caption{Test accuracy on MvMidog. In this set of experiments, we set the upload ratio of the statistics r = 0.1. To ensure fairness, each algorithm is evaluated three times, and the final average is taken as the experimental result. } 
\label{tab:mvmidog} 
\begin{tabular}{lccccc|ccccc}
\hline
& \multicolumn{10}{c}{\textbf{MvMidog}} \\ 

& \multicolumn{5}{c|}{\textbf{Resnet18}} & \multicolumn{5}{c}{\textbf{Resnet50}} \\ 

& H & C & S & X & Avg. & H & C & S & X & Avg. \\ \hline
FedAvg & 76.15 & 72.11 & 88.77 & 90.14 & 81.79 & 77.06 & 81.09 & 88.77 & 81.43 & 81.67 \\
FedProx & 81.2 & \textbf{87.54} & 89.02 & 82.85 & 85.15 & 81.96 & 85.64 & 89.02 & 90.73 & 86.88 \\
FedCCRL & 79.41 & 84.00 & 89.23 & 91.82 & 86.11 & 82.21 & 85.99 & 89.23 & 88.48 & 86.38 \\
FedAlign & 78.65 & 77.95 & 87.29 & 88.61 & 83.13 & \textbf{82.32} & 78.01 & 86.98 & 87.04 & 83.59 \\
Strap & 71.30 & 80.00 & 86.67 & 90.00 & 81.99 & 66.67 & 73.33 & 73.65 & 83.67 & 74.33 \\
FedStain(ours) & \textbf{81.23} & 83.10 & \textbf{91.29} & \textbf{91.27} & \textbf{86.72} & 81.05 & \textbf{88.11} & \textbf{92.26} & \textbf{90.82} & \textbf{88.06} \\ \hline
\end{tabular}
\end{table*}
We benchmarked FedStain against both conventional federated learning algorithms (e.g., FedAvg, FedProx) and state-of-the-art FDG approaches (e.g., FedCCRL, Strap). Following the partitioning strategy outlined in Section 3.1, we evaluated all methods under varying numbers of clients per domain to simulate realistic federated scenarios.The backbone network adopted in all experiments was ResNet50, where the early convolutional layers functioned as a feature encoder and the final fully connected layer acted as a  task-specific classifier. To improve stain and texture invariance during pretraining, we incorporated the RandStainNA augmentation strategy (probability p=0.9), which perturbs color and contrast distributions to emulate diverse staining conditions.Building upon the MixStyle-based inter-client statistics exchange employed in FedCCRL, FedStain extends the representation by introducing higher-order statistics—specifically, skewness and kurtosis—enabling richer cross-domain information flow beyond mean–variance statistics.\par
Hyperparameter optimization was carefully performed to ensure stable convergence. The selected parameters, summarized in Tables \ref{tab:single} and \ref{tab:single1}, align with established practices in federated generalization research. All experiments utilized the Adam optimizer with a temperature parameter $\tau = 0.1$. Input patches were resized to 128×128, and a cosine learning rate schedule was applied to achieve smooth decay across training epochs. Both the RandStainNA and MixStyle modules were actively exclusively during local client training, maintaining consistent configurations throughout experiments to guarantee reproducibility and fairness.\par
To account for stochastic variability, each experiment was independently repeated three times on different computational setups, and the mean performance is reported.

\subsection{Comparison with Baselines}
We conducted comprehensive comparisons on two histopathology benchmarks, Camelyon17 and MvMidog, against a variety of representative federated learning approaches, including classical optimization-based methods (FedAvg, FedProx) and state-of-the-art domain-generalization-oriented algorithms (FedCCRL, FedAlign, Strap). All methods were evaluated under the same configuration with a statistic upload ratio of r=0.1, and each experiment was repeated three times to report the averaged results. Both ResNet18 and ResNet50 backbones are considered. Complete quantitative results are provided in Tables \ref{tab:camelyon} and \ref{tab:mvmidog}.\par

1) Camelyon17: Across the five test hospitals, FedStain consistently achieves the highest overall accuracy under both ResNet18 and ResNet50. Compared with both classical federated baselines (FedAvg, FedProx) and domain-generalization methods (FedCCRL, FedAlign, Strap), FedStain shows substantial and stable improvements, demonstrating a markedly stronger robustness to inter-hospital domain shifts. A particularly notable observation lies in Hospital 4, the most challenging domain, characterized by severe stain deviations, scanner differences, and a clearly non-Gaussian stain distribution. Under such high-order distributional shifts, most prior methods experience substantial performance degradation—for example, FedCCRL and FedAlign drop to 69.47\% and 79.34\%, respectively—highlighting their limited robustness to asymmetric stain variations. In contrast, FedStain explicitly exchanges higher-order statistics (skewness and kurtosis), enabling the model to capture complex non-Gaussian stain patterns. As a result, FedStain attains 95.01\% accuracy on Hospital 4, outperforming all competing methods by a large margin and maintaining accuracy comparable to that of other hospitals.\par

2) MvMidog: MvMidog introduces even larger cross-domain discrepancies due to differences in tissue structures, staining protocols, and imaging devices. Under these highly heterogeneous conditions, FedStain continues to demonstrate strong domain generalization. Using ResNet50 as the backbone, FedStain achieves the highest mean accuracy of 88.06\%, outperforming all baselines and confirming its robustness when facing multi-source distribution shifts.\par

Overall, across two datasets with substantial domain variability and two backbone architectures, FedStain consistently delivers the best average performance by significant margins. Compared with prior methods, FedStain exhibits superior cross-domain stability and robustness. The effectiveness of exchanging higher-order statistics, combined with stain-aware augmentation, is thoroughly validated through experiments which showing clear advantages in mitigating domain shifts induced by stain variations and scanner discrepancies.

\subsection{Ablation Study}
\begin{table}[h] 
\centering
\caption{Accuracy on the MvMidog dataset using different statistical features.}
\label{tab:mvmidog-features}
\newcolumntype{C}[1]{>{\centering\arraybackslash}m{#1}}
\renewcommand{\tabularxcolumn}[1]{m{#1}} 
\begin{tabularx}{\columnwidth}{Xccccc}
\hline
& \multicolumn{5}{c}{\textbf{MvMidog}} \\
& H & C & S & X & Avg. \\ \hline
Mean \& Std & 76.25 & 82.00 & 90.59 & 89.78 & 84.66 \\
Channel 90\% quantile \& Compilation coefficient & 76.88 & 80.87 & 89.05 & 90.80 & 84.40 \\
Local mean \& Local MAD & 74.18 & 84.26 & 86.59 & 90.82 & 83.96 \\
Mean \& Interquartile range & 74.16 & 78.16 & 88.77 & 88.35 & 82.36 \\
Skewness \& Kurtosis & \textbf{81.05} & \textbf{88.11} & \textbf{92.26} & \textbf{90.82} & \textbf{88.06} \\ \hline
\end{tabularx}
\end{table}

To further analyze the contribution of statistical components, we conducted ablation experiments involving different combinations of exchanged statistics. Specifically, we compared five statistical pairs:\par
1) mean \& std,\par
2) channel 90\% quantile \& channel compilation coefficient,\par
3) local mean \& local MAD,\par
4) mean \& interquartile range (IQR), \par
5) skewness \& kurtosis,\par
\begin{figure*}[htbp]
    \centering
    \begin{subfigure}[b]{0.48\textwidth}
        \centering
        \includegraphics[width=\linewidth]{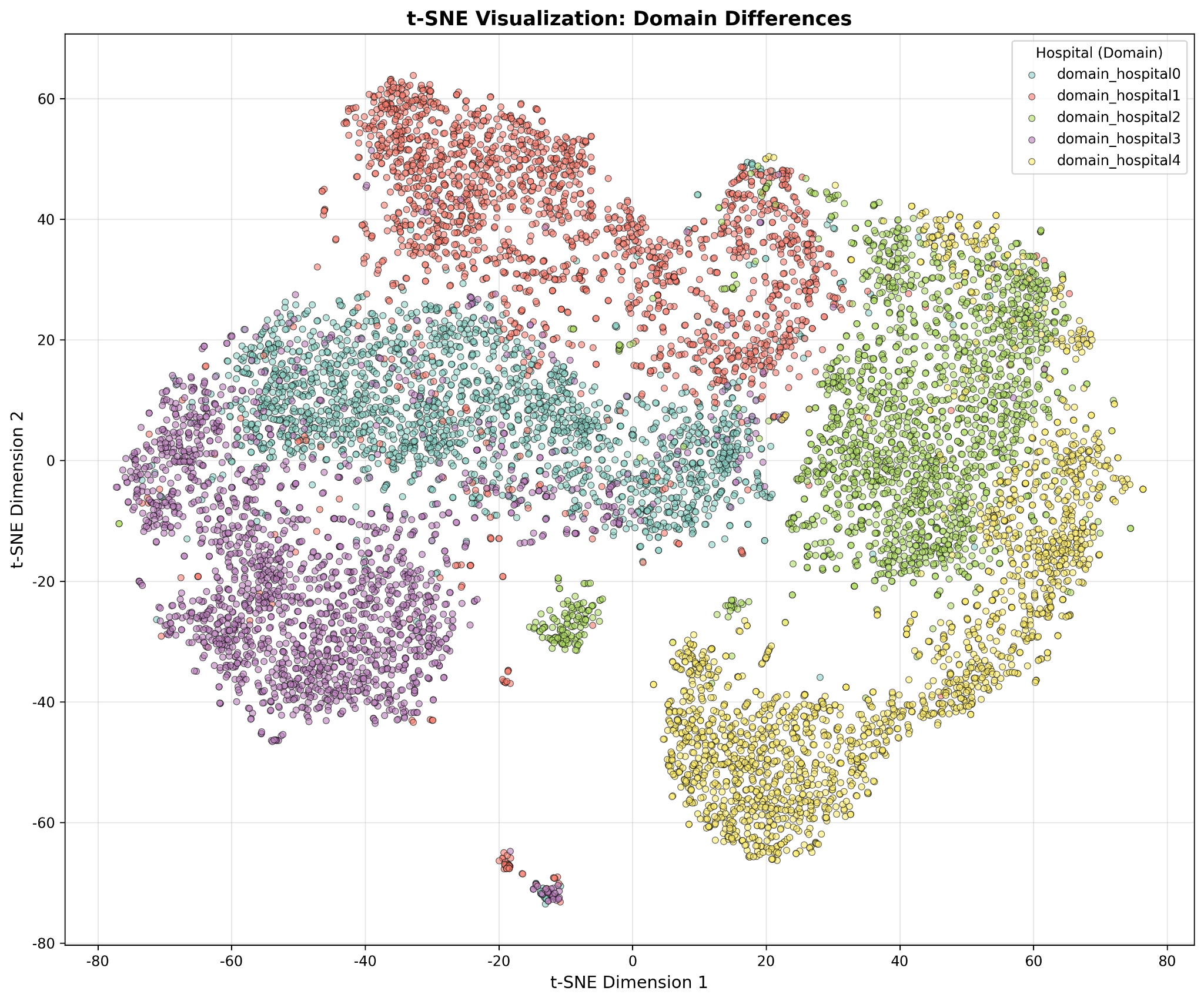} 
        \caption{Camelyon17}
        \label{fig:left_img}
    \end{subfigure}
    \hfill 
    \begin{subfigure}[b]{0.48\textwidth}
        \centering
        \includegraphics[width=\linewidth]{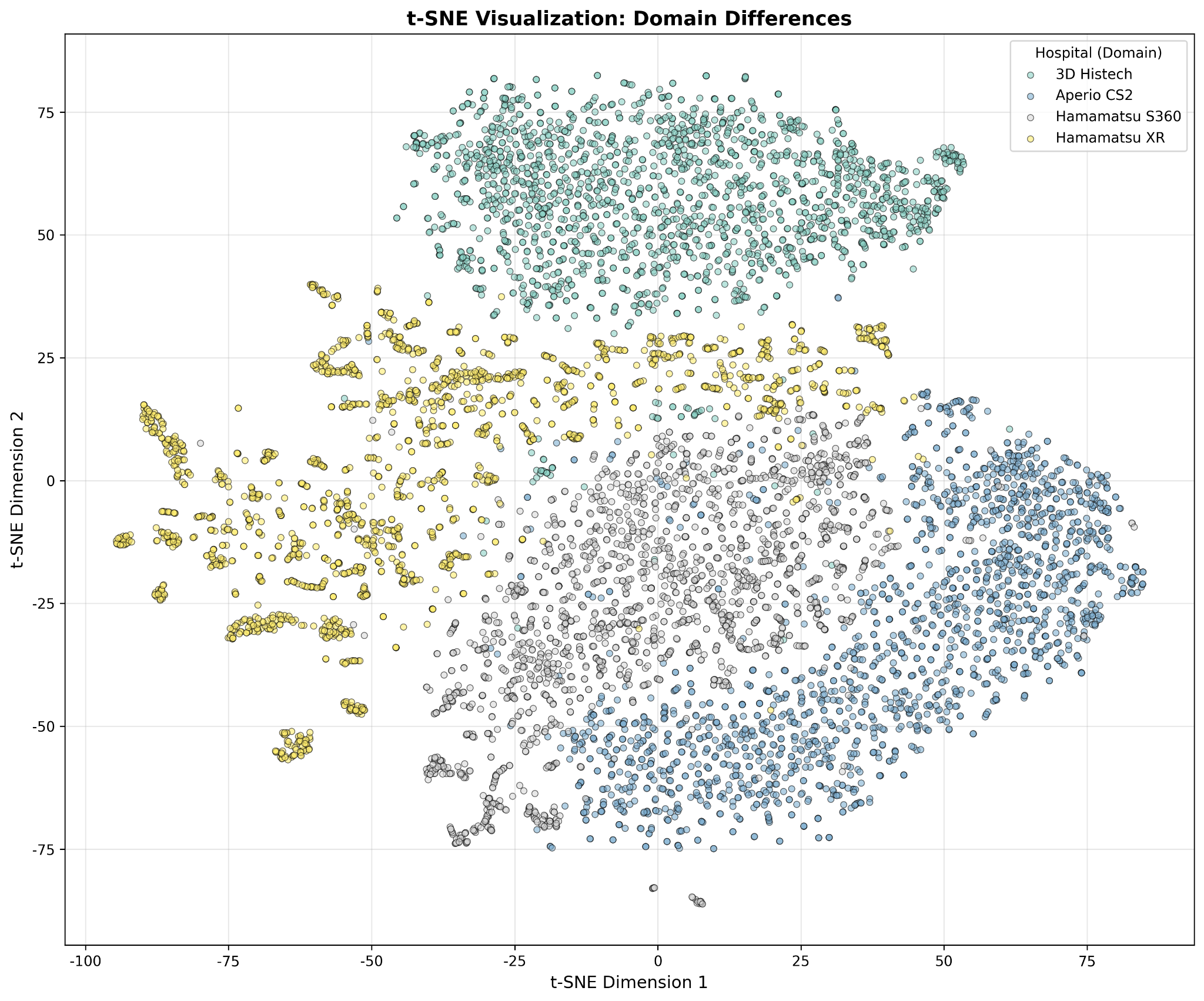}
        \caption{MvMidog}
        \label{fig:right_img}
    \end{subfigure}
    
    \caption{The t-SNE visualization of histopathology embeddings. 
    Results reveal substantial domain gaps across (a) hospitals in Camelyon17 and (b) scanners in MvMidog, illustrating strong inter-domain separability.}
    \label{fig:two_images}
\end{figure*}
The results (Table V) clearly demonstrate that the combination of skewness and kurtosis consistently achieves the most robust performance across all domains. Notably, our proposed configuration surpassed the baseline FedCCRL, which exchanges only mean and variance, in every evaluation setting. This suggests that relying solely on low-order statistics may not fully characterize the complex and non-Gaussian stain distributions commonly observed in real clinical settings. In contrast, higher-order statistics can capture richer structural cues related to the underlying distributional shape, enabling more effective alignment of asymmetric and long-tailed stain variations across clients.

\subsection{Feature Distribution Visualization}
1)t-SNE: To further examine the intrinsic domain shift present in histopathology datasets, we projected the raw feature embeddings onto a 2D space using t-SNE, as shown in Fig.2. In the Camelyon17 dataset, the samples originating from different hospitals form clearly separated clusters with sharp boundaries, reflecting substantial inter-institutional staining variability. A similar pattern is observed in the MvMidog dataset, where patches scanned by different whole-slide imaging devices exhibit pronounced cluster structures, indicating distinct stain appearance and scanner-specific characteristics. The strong separability across domains in both datasets highlights the severity of domain shift and validates the necessity of explicitly modeling domain-aware statistical properties.\par
\begin{figure}[t]
\centering
\begin{tabular}{ >{\raggedleft\arraybackslash}m{1.5cm} c c c } 
& \textbf{FedAvg} & \textbf{FedCCRL} & \textbf{FedStain} \\ [1ex]
Camelyon17 &
\includegraphics[width=2cm, valign=c]{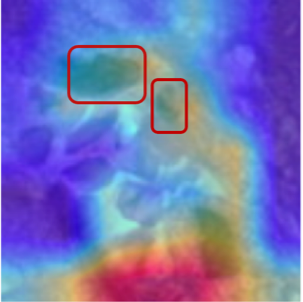} &
\includegraphics[width=2cm, valign=c]{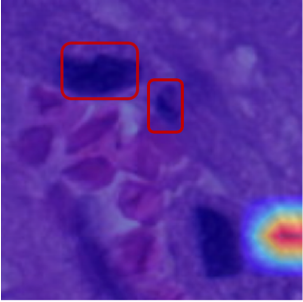} &
\includegraphics[width=2cm, valign=c]{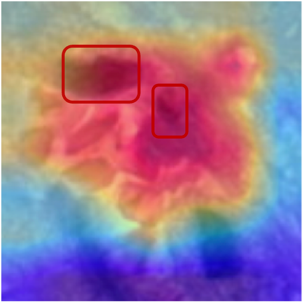} \\ [1ex]
MvMidog &
\includegraphics[width=2cm, valign=c]{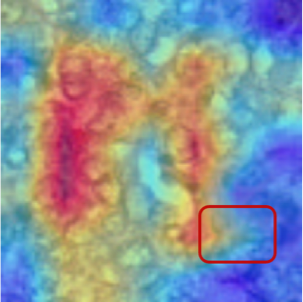} &
\includegraphics[width=2cm, valign=c]{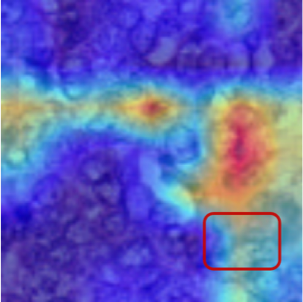} &
\includegraphics[width=2cm, valign=c]{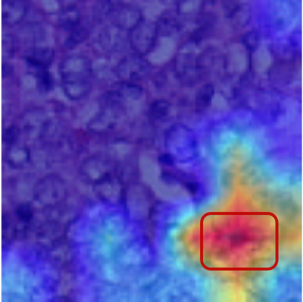}
\end{tabular} 
\caption{ Grad-CAM visualizations on Camelyon17 and MvMidog.Red boxes mark cancerous regions.FedAvg and FedCCRL show dispersed, background-biased attention, while FedStain focuses precisely on lesion areas, indicating stronger domain-invariant localization.}
\label{fig:vis-results}
\end{figure}
2)Grad-CAM: To further assess whether the learned representations are domain-invariant, Figure 3 presents the Grad-CAM visualizations for different methods. The red annotations highlight the precise locations of the cancerous tissue.A comparative analysis reveals significant differences in attention mechanisms. The heatmaps generated by FedAvg and FedCCRL exhibit scattered attention, often highlighting non-discriminative background areas outside the annotated regions. This indicates that the baselines are susceptible to domain shifts, causing the model to overfit to stain styles rather than the object of interest. Conversely, our proposed FedStain suppresses such background interference and concentrates its attention strictly within the ground-truth boundaries. By accurately focusing on the cancer regions despite stain variations, FedStain proves its efficacy in extracting robust, domain-invariant semantic features, which is critical for reliable federated diagnosis.\par

\section{Conclusion}
In this paper, we introduced FedStain, a novel FedDG framework tailored for medical image analysis. FedStain significantly improves model generalizability to data from unseen hospitals or scanners without violating privacy regulations or introducing excessive computational and communication overhead.\par
Specifically, FedStain strategically integrates higher-order statistics to effectively achieve cross-site style transfer and perturbation of domain-invariant features. Furthermore, we leverage a Supervised Contrastive Loss to facilitate robust representation alignment and apply the Jensen-Shannon Divergence for prediction alignment across clients.\par
Extensive experiments conducted on two representative medical benchmarks, Camelyon17 and Mvmidog, demonstrate that FedStain outperforms state-of-the-art baselines, thereby validating its superior effectiveness and robustness in medical image classification tasks under the challenging FedDG setting.

\end{document}